\documentclass[preprint,twocolumn,5p,12pt]{article}

\usepackage{amssymb}
\usepackage{subcaption}
\usepackage{graphicx}
\usepackage{caption}
\usepackage{xspace}
\usepackage{multirow}
\usepackage{rotating}
 \usepackage{url}
\usepackage{gensymb}
\usepackage [ table ]{ xcolor }
\usepackage[top=2cm, bottom=2cm, left=1.5cm, right=1.5cm]{geometry}
\usepackage{setspace}
\usepackage{morefloats}

\usepackage{commath}

\usepackage[affil-it]{authblk}
\usepackage[colorlinks=true,citecolor=blue, urlcolor=blue, linkcolor=blue]{hyperref}

\title{Spiking neural networks trained via proxy}

\author{Saeed Reza Kheradpisheh$ ^{1,}$\footnote{Corresponding Author \\Email addresses:\\ \href{mailto://s_kheradpisheh@sbu.ac.ir}{s\_kheradpisheh@sbu.ac.ir} (FA), \\ \href{mailto://mirsadeghi@aut.ac.ir}{mirsadeghi@aut.ac.ir} (SA), \\ \href{mailto:/timothee.masquelier@cnrs.fr}{timothee.masquelier@cnrs.fr} (TA)} }
\author{Maryam Mirsadeghi$ ^{2}$}
\author{Timoth\'ee Masquelier$ ^{3}$}
\affil{\footnotesize $ ^{1} $ Department of Computer Science, Faculty of Mathematical Sciences, Shahid Beheshti University, Tehran, Iran}
\affil{\footnotesize $ ^{2} $  Department of Electrical Engineering, Amirkabir University of Technology, Tehran, Iran}
\affil{\footnotesize $ ^{3} $  CerCo UMR 5549, CNRS Universit\'e Toulouse 3, France}

\date{}

\usepackage[absolute]{textpos}
\begin{document}
\begin{textblock}{19}(1,1)
\noindent \textbf{\color{red} This manuscript is published in \textbf{IEEE Access}. Please cite it as:}\\
\textit{\color{blue} S.R. Kheradpisheh, M. Mirsadeghi, T. Masquelier,  Spiking neural networks trained via proxy. IEEE Access 10 (3187033) (2022)  70769-70778\\doi: \url{https://doi.org/10.1109/ACCESS.2022.3187033}}
\end{textblock}

\maketitle

\begin{abstract}
We propose a new learning algorithm to train spiking neural networks (SNN) using conventional artificial neural networks (ANN) as proxy. We couple two SNN and ANN networks, respectively, made of integrate-and-fire (IF) and ReLU neurons with the same network architectures and shared synaptic weights. The forward passes of the two networks are totally independent. By assuming IF neuron with rate-coding as an approximation of ReLU, we backpropagate the error of the SNN in the proxy ANN to update the shared weights, simply by replacing the ANN final output with that of the SNN. We applied the proposed proxy learning to deep convolutional SNNs and evaluated it on two benchmarked datasets of Fashion-MNIST and Cifar10 with 94.56\% and 93.11\% classification accuracy, respectively. The proposed networks could outperform other deep SNNs trained with tandem learning, surrogate gradient learning, or converted from deep ANNs. Converted SNNs require long simulation times to reach reasonable accuracies while our proxy learning leads to efficient SNNs with much smaller simulation times. The source codes of the proposed method are publicly available at \href{https://github.com/SRKH/ProxyLearning}{https://github.com/SRKH/ProxyLearning}.
\end{abstract}
\section{Introduction}
Artificial intelligence has been revolutionized by the successes of deep learning during the past decade. However, the top performing deep learning models consume a huge amount of power and computation to run and be trained\cite{thompson2020computational}. For instance, training of the  GPT3 language model requires 190,000 $kWh$, while our whole brain consumes around 12-20 $W$ of power to work~\cite{eshraghian2021training}. Spiking neural networks (SNNs),  as the 3rd generation of neural networks, are largely inspired by the neural computations in the brain. SNNs are known to be suitable for hardware implementation specially on neuromorphic devices~\cite{zenke2021visualizing,bouvier2019spiking}, which are quite fast and require much less amount of power. However, the temporal dynamics of SNNs both in neuron and network levels along with the non-differentiabilty of spike functions have made it difficult to train efficient SNNs~\cite{tavanaei2019deep}. Different studies with different approaches have tried to adapt backpropagation based supervised learning algorithms to SNNs~\cite{pfeiffer2018deep}. 

The first approach is to train an artificial neural network (ANN) and then convert it to an equivalent SNN~\cite{rueckauer2018conversion,rathi2020enabling,sengupta2019going,10.3389/fnins.2020.00119,deng2021optimal}. Although converted SNNs could be applied to deep architectures and reached reasonable accuracies, they  totally neglect the temporal nature of SNNs and usually are inefficient in terms of the number of spikes/time-steps. The second approach is to directly apply backpropagation on SNNs. Their main challenge is to overcome the non-differentiablity  of spike functions required in  error-backpropagation algorithm. To solve this problem, some studies propose to use smoothed spike functions with true gradients~\cite{huh2018gradient} and others use surrogate gradients for non-differentiable discrete spike functions~\cite{neftci2019surrogate,bohte2011error,esser2016convolutional,shrestha2018slayer,bellec2018long,zimmer2019technical,pellegrini2021low,pellegrini2021fast}. The main issue with this approach is the use of backpropagation through time which makes it too costly and faces it with vanishing/exploding gradient problem, especially for longer simulation times. 

In the third approach, known as latency learning, the idea is to define the neuron activity as a function of its firing time~\cite{kheradpisheh2020temporal,zhang2020spike,sakemi2021supervised,zhang2020temporal,bohte2002error,zhou2019direct,wunderlich2021event}. In other words, neurons fire at most once and stronger outputs correspond to shorter spike delays. To apply backpropagation to such SNNs, one should define the neuron firing time as a differentiable function of its membrane potential or the firing times of its afferents~\cite{mostafa2017supervised,kheradpisheh2020temporal}. As an advantage, latency learning  does not need backpropagation through time, however, it is more difficult to train outperforming SNNs with latency learning.

The forth approach is the tandem learning which is consists of an SNN and an ANN coupled layer-wise through weight sharing~\cite{wu2021tandem,wu2020progressive}. The auxiliary ANN is used to facilitate the error backpropagation for the training of the SNN at the spike-train level, while the SNN is used to derive the exact spiking neural representation. Literally, in the forward pass,  each ANN layer receives its input as the spike counts of the previous SNN layer, and consequently, in the backward pass, each ANN layer computes the gradient of its output with respect to the shared weights based on these input spike counts. Regarding this layer-wised coupling, the learning can be also done in a progressive layer by layer manner.

In this paper, we propose a new learning method to train an SNN via a proxy ANN. To do so, we make an ANN (consists of ReLU neurons) structurally equivalent to the SNN (made of integrate-and-fire (IF) neurons). The two network share their synaptic weights, however, IF neurons in the SNN work with temporally distributed spikes, while, neurons in ANN work with floating points and process their input instantly.  By considering IF with rate coding as an approximation of ReLU, we replace the final output of the SNN into that of the ANN, and therefore, we update the shared weights by backpropagating the SNN error in the ANN model. In other words, we assume that greater output values in ReLU neurons is approximated with higher firing rates in the equivalent IF neurons. Hence, the SNN error is an approximation of ANN error and by backpropagating the SNN error in the proxy ANN, the shared weights are updated according to the SNN error but based on the differentiable  activations of the ANN.

One of the main challenges in conversion and tandem learning methods is the approximation of the ANN max-pooling layers in the corresponding SNN, hence,  they  exclude pooling layers and use convolutional layers with stride of 2. Here, we used spike-pooling layers to mimic the behavior of max-pooling layers in the corresponding ANN. We evaluated the proposed proxy learning on Cifar10 and Fashion-MNIST dataset with different deep convolutional architectures and outperformed the currently existing conversion and tandem learning methods.

The main advantages of the proposed proxy learning method with respect to the previous learning methods are:
\begin{itemize}
    \item Contrary to conversion methods, we make the backpropagation in  ANN aware of the error of the SNN by replacing the ANN output with  that of the SNN.
    \item Contrary to tandem learning, during the learning, the forward pass of the two networks are totally independent with no interference.
    \item Contrary to the surrogate gradient learning, we do not need to backpropagate the SNN error in time which is computationally expensive, memory consuming, and suffering from gradient vanishing/exploding problem. 
\end{itemize}
\section{Spiking neural networks}
Inspired by the brain, spiking neural networks (SNNs) use spikes to compute, transmit, and learn information. Therefore, running SNNs on hardware or neuromorphic devices would require much less amount of power and computation with respect to traditional ANNs~\cite{eshraghian2021training}. Because of that, SNNs has recently been widely attended and extended by scientific researchers and industrial companies. Also SNNs work in temporal domain, which makes them a better solution for learning and processing of spatiotemporal patterns~\cite{tavanaei2019deep}. Despite the aforementioned advantages of SNNs, they are yet behind ANNs in terms of accuracy. One of the main challenges for developing powerful SNNs with low computational and energy cost is the lack of efficient supervised learning methods. Applying backpropagation to SNNs is not straightforward, as the spiking activation functions are not differentiable. Also, SNNs works in time, and the error should be backpropagated in time which requires a lot of memory and time and also it suffers from vanishing/exploding gradient problem.   
 
 Different solutions are proposed to tackle these issues including the use of soft and differentiable spike functions~\cite{neftci2019surrogate}, using surrogate gradients for impulse spike functions~\cite{neftci2019surrogate,fang2020incorporating}, converting trained ANNs to equivalent SNNs~\cite{rueckauer2017conversion,rathi2020enabling}, and training SNN layers using ANN layers in tandem~\cite{wu2021tandem,wu2020progressive}. However none of these solutions can solve all the aforementioned issues with applying backpropagation to SNNs. Here, we propose the proxy learning method which simply backpropagates the SNN error in an equivalent proxy ANN (with shared weights) with no need for backpropagation in time or the use of surrogate gradients. Contrary to conversion methods, in proxy learning, the ANN is aware of the SNN error, and contrary to tandem learning methods, the forward passes of the two ANN and SNN models are completely independent. In the proposed model, we use the non-leaky integrate-and-fire (IF) neuron model which has the simplest neural dynamics amongst all the spiking neuron models. 
 
 In the following subsections, we explain different aspect of the proposed SNN model including the IF neuron model (Subsection~\ref{IFNeuron}), the direct input encoding scheme (Subsection~\ref{encoding_section}), and convolutional SNN architecture (Subsection~\ref{convolution_section}). Then, in Section~\ref{IFrelu}, we illustrate how the ReLU neuron model can be approximated with IF neuron model and rate coding, and finally, in Section~\ref{TPSection}, we explain the inference (forward pass) and  the learning (backward pass) of the proposed model.

\subsection{Spiking neuron}\label{IFNeuron}
Artificial neurons are simply implemented by a linear combination of inputs followed by a non-linear activation function. Different activation functions have different mathematical properties and choosing the right function can largely impact the learning speed and efficiency of the whole network. The revolution of deep learning was accompanied by the use of Rectified Linear Unit (ReLU) instead of prior activation functions such as sigmoid and hyperbolic tangent. An artificial neuron $j$ with ReLU activation can be formulated as follows,
\begin{equation}\label{Eq00}
    z_j=\sum_i w_{ji}x_i, 
\end{equation}
\begin{equation}\label{Eq01}
    y_j=ReLU(z)=max(0,z), 
\end{equation}
where $x$ and $w$ are respectively the input and weight vectors, and $y_j$ is the neuron output. 

Contrary to the artificial neuron models which work with synchronous  instant inputs, spiking neurons have a temporal dynamics through which the neuron's internal membrane potential changes in time by receiving asynchronous incoming spikes. The complexity of this neural dynamics can largely impact the computational cost and learning efficiency of SNNs. Integrate-and-fire (IF) is the simplest spiking neuron model in which the membrane potential only changes when receiving an input spike from a presynaptic neuron by an amount proportional to the synaptic weight. 

The membrane potential $u_j$ of an IF neuron $j$ is updated at each time step $t$ by the input current $I_j(t)$ caused by the spike train $s_i(t)$ received from each presynaptice neuron $i$ through a synapse with the weight $w_{ji}$,
\begin{equation}\label{Eq1}
    u_j(t)= u_j(t-1)+ RI_j(t), 
\end{equation}
\begin{equation}\label{Eq2}
    I_j(t)= \sum_i w_{ji}s_i(t), 
\end{equation}
where $s_i(t)=1$ if presynaptic neuron $i$ has fired at time $t$ and it is zero elsewhere. We set the membrane resistance $R$ to unitary (i. e., $R=1$).

The IF neuron emits an output spike whenever its membrane potential crosses a certain threshold $\theta$,
\begin{equation}\label{Eq3}
    s_j(t)=
    \begin{cases}
    1 & \quad \text{if  } V_j(t) \geq \theta,\\
    0 & \quad \mathrm{otherwise},
    \end{cases}
\end{equation}
and then resets its membrane potential to zero as $u_j=0$ to be ready for the forthcoming input spikes.
\subsection{Information encoding}\label{encoding_section}
ANNs work with decimal values, therefore, their inputs are represented by vectors, matrices, and tensors of floating-point numbers. However, in SNNs, neurons talk through spikes, hence, information needs to be somehow encoded in the spike trains. In other words, the analog input signal should be converted into an equivalent spike train in the entry layer of the network. Different coding schemes are suggested to be used in SNNs ranging from heavy rate codes to economical temporal codes with single spikes. 

Another spike-free input coding approach is to use constant input currents (aka direct input coding) applied to input neurons. This way, input neurons with higher input currents will fire with higher rates than others. In other words, contrary to the neurons in the middle and output layers, the input current to the IF neurons in the first layer is proportional to the input signal. Consider an input image $x$, the constant input current to an input IF neuron $j$ is computed as
\begin{equation}\label{Eq4}
    I_j(t)= \sum_i w_{ji}x_i, 
\end{equation}
where $x_i$ is the $i^{th}$ input pixels inside the receptive field of neuron $j$ and $w_{ji}x_i$ is the constant input current from $x_i$ to  $j$.  
\subsection{Convolutional SNN}\label{convolution_section}
Convolutional ANNs (CANNs), largely inspired by the hierarchical object recognition process in visual cortex, are comprised of a cascade of interlaying convolution and pooling layers to extract descriptive visual features followed by a stack of fully connected layers to make the final decision. Neural processing in  CANNs are performed in a layer by layer fashion, neurons in each layer receive their whole input from the previous layer at once, and instantly send their output to the next layer. The neural processing in convolutional SNNs (CSNNs) is different as neurons work in temporal domain and information is encoded in asynchronous spike trains.

Each convolutional layer of a CSNN is made of IF neurons organized in numerous feature maps. At every time step, each neuron receives the total input current from the afferents inside its receptive window (Eq.~\ref{Eq4} for input layer and Eq.~\ref{Eq2} for other layers), updates its membrane potential (Eq.~\ref{Eq1}) and fires whenever reaches to the threshold (Eq.~\ref{Eq3}). Neurons in the same feature map share their synaptic weights, and hence, look for the same feature but in different locations. 

At each time step, neurons in spike-pooling layers of the CSNN simply emit an spike if there is at least one spike in their input window. Hence, if different neurons inside the receptive window of a spike-pooling neuron fire at several different times, the spike-pooling neuron will also fire at each of those time steps. Spike-pooling neurons can simply be implemented by IF neurons with synaptic weights and threshold of one.

Fully-connected layers of the CSNN, including the readout layer, are all made of IF neurons. The  last convolution or spike-pooling layer is flattened and its spikes are fed to the first fully-connected layer. The readout layer includes one neuron for each class, and the neuron with the maximum number of spikes determines the winner class.
\section{IF approximating ReLU}\label{IFrelu}
Although the input encoding, internal mechanism, and output decoding of IF are different from those of ReLU neuron model, several studies from different perspectives have shown that an equivalent IF (or Leaky-IF) neuron  can fairly approximate the activation of the ReLU neuron. For time-to-first-spike coding, the firing time of  IF neuron is inversely proportional to the output of ReLU ~\cite{kheradpisheh2020temporal,rueckauer2018conversion}. In short, the IF neuron remains silent if the output of ReLU is zero,  and it will fire with shorter delay for larger ReLU outputs. For rate coding, the higher ReLU outputs  correspond to higher firing rates in IF neuron~\cite{wu2021tandem,wu2020progressive,tavanaei2019bp}.

Here, we consider that neural information is encoded in the spike rate of neurons. Let $r_i$ be the input spike rate received from the $i^{th}$ afferent to neuron $j$, 
\begin{equation}
    r_i=\frac{\sum_t s_i(t)}{T},
\end{equation}
where $T$ is the maximum simulation time. For simplicity, we assume that these input spikes are uniformly distributed in time and the sudden current caused by each spike is evenly and continuously delivered during the inter-spike intervals, then according to Eq.~\ref{Eq2}, the input current to neuron $j$ is constant in time and is calculated as,  
\begin{equation}
    I_j=\sum_i w_{ji}r_i.
\end{equation}

If we apply this constant input current to IF neuron in Eq.~\ref{Eq1} with the thresholding function in Eq.~\ref{Eq3}, the firing rate of IF neuron $j$ can be calculated as,
\begin{equation}\label{Eq9}
r_j=ReLU(\frac{RI_j}{\theta})=\frac{R}{\theta}\:ReLU(\sum_i w_{ji}r_i).
\end{equation}
The use of ReLU is necessary here, since the IF neuron will not fire at all for negative input currents. As shown in Eq.~\ref{Eq9}, the firing rate of the IF neuron $j$ can be expressed by applying ReLU on the weighted summation of the input firing rates from its afferents.

\begin{figure}[t]
    \begin{center}
    \includegraphics[width=.32\textwidth]{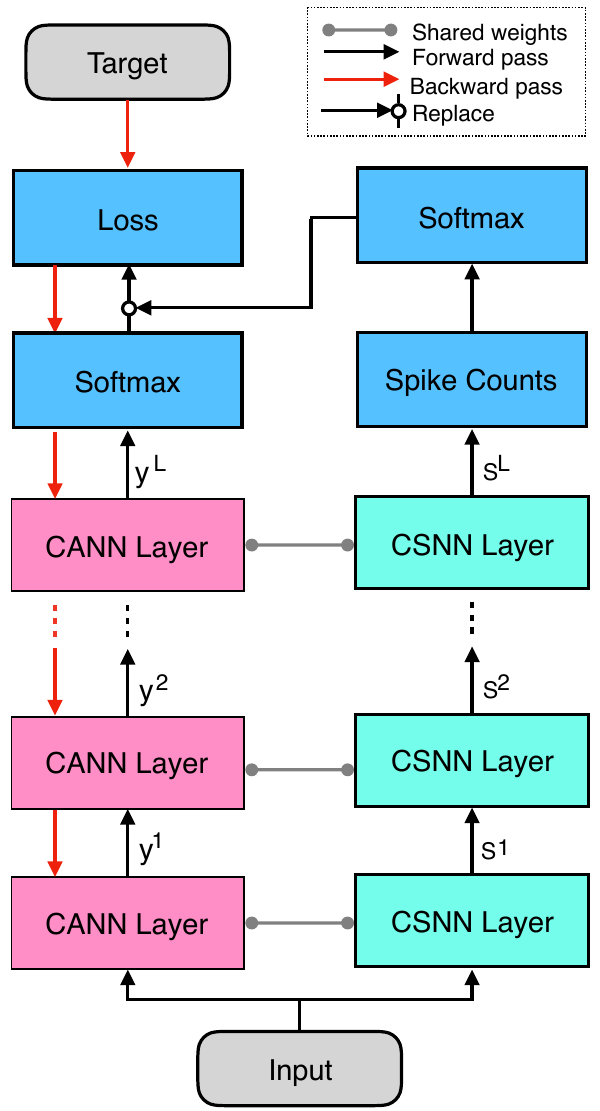}
    \caption{The proposed proxy learning that is comprised of a CSNN coupled with a CANN through the shared weights. The $s^l$ and $y^l$ are respectively the spike train and the output of the $l^{th}$ CSNN and CANN layers during the forward pass. The final output of the CANN is replaced with the output of the CSNN model. During the Backward pass, the error of the CSNN is backpropagated through the CANN network to update the shared weights.}  
    \label{fig1}
    \end{center}
\end{figure}
\section{Training via proxy}\label{TPSection}
The general idea of the proposed learning method is illustrated in Fig.~\ref{fig1}. Here we have a proxy CANN coupled with an equivalent CSNN having same architectures with $L$ layers. The weights in all layers are shared between the two networks. However, the CANN model is made of artificial neurons with ReLU activation and the CSNN model is made of IF neurons and works in the temporal domain. The input image is fed to both model and their outputs are obtained at their softmax layer. Since the thresholding function of IF neurons is not differentiable, it can not be directly trained. We discard the output of CANN (the proxy network) and replace it with the CSNN output, then, we backpropagate the CSNN error in CANN to update the the weights.

\subsection{Forward pass}
During the forward pass the input image is fed to both networks. The first layer of CANN simply convolves its filter kernels over the input image and sends it output, $y^1$ to the next layer by applying the ReLU activation function. The following pooling and convolutional layers apply the max-pooling and convolution operations on their inputs from the previous layer and send their output to the next layer. The fully-connected layers on top receive inputs from all neurons in their previous layer through synaptic weights and send their output to the next layer. Note that we use ReLU activation function in all convolutional and fully-connected layers. At the end, CANN applies softmax activation function over the output of the last layer, $y^L$, to obtain the final output, $O^A$,
\begin{equation}
    O_k^A= \frac{e^{y_k^L}}{\sum_j e^{y_j^L}}.
\end{equation}

The first layer of the CSNN model obeys the input encoding scheme explained in Section~\ref{encoding_section}. The CSNN process each input image in $T$ simulation time step. At every time step, the constant input current to IF neurons in the first layer of the CSNN model is computed by convolving the input image with the corresponding filter kernel (see Eq.~\ref{Eq4}). These IF neurons will emit spikes whenever their membrane potential crosses the threshold and send spike train $s^1$ to the next layer. As explained in Section~\ref{convolution_section}, in any time step, spike-pooling neurons will fire a spike if there is at least one spike in their receptive window. Spiking convolutional IF neurons integrate incoming spikes inside their receptive window through weighted synapses and fire when the threshold is reached. It is the same for IF neurons in fully-connected layers but they do not have restricted receptive window and receive spikes from all neurons in the previous layer. To compute the output of CSNN, we count the number of spikes for each neuron in the output layer, $C_k^L$, and send it to a softmax layer to obtain final output, $O^S$,

\begin{equation}
    c_k^L= \sum_t s_k^L(t),
\end{equation}
\begin{equation}
    O_k^S= \frac{e^{c_k^L}}{\sum_j e^{c_j^L}}.
\end{equation}

\subsection{Backward pass}
As mentioned earlier, to update the weights of the CSNN model we use the corresponding shared weights in CANN. To do so, the softmax output of the CANN model is replaced by the  softmax output of the CSNN model. By comparing it to the target values in the loss function of the CANN model, we are literally computing the error and the loss of the CSNN model. This loss is then backpropagated through the CANN model to update the shared weights. As explained in Section~\ref{IFrelu}, we assume that the input and output of the ReLU neuron in the CANN model is approximated by the input and output firing rates of the corresponding IF neuron in the CSNN model

Let assume that $Y_k$ is the target of the $k^{th}$ output in the CANN model. The cross-entropy loss function for the CANN model is defined as
\begin{equation}\label{eq13}
    L=-\sum_k Y_k ln(O^A_k) \simeq -\sum_k Y_k ln(O^S_k). 
\end{equation}

To update  an arbitrary shared weight $w^l_{ji}$ in the $l^{th}$ layer through the CANN model, we have 
\begin{equation}
\Delta w^l_{ji}=w^l_{ji} -\eta\frac{\partial L}{\partial w^l_{ji}}
\end{equation}
where $\eta$ is the learning rate. Instead of using the true gradient of the CANN model which is computed as
\begin{equation}\label{bpANN}
    \frac{\partial L}{\partial w^l_{ji}}=\sum_k \frac{\partial L}{\partial O^A_k}\sum_d \frac{\partial O^A_k}{\partial y_d^L} \frac{\partial y_d^L}{\partial w^l_{ij}},
\end{equation}
we use the following approximated gradient which is obtained by replacing the output of the CSNN into the output of the CANN model,
\begin{equation}\label{bpSNN}
    \frac{\partial L}{\partial w^l_{ji}} \simeq \sum_k \frac{\partial L}{\partial O^S_k}\sum_d \frac{\partial O^S_k}{\partial y_d^L} \frac{\partial y_d^L}{\partial w^l_{ij}}.
\end{equation}

Indeed, we are backpropagating the error of the CSNN model in the CANN model to update the shared weights. Literally, in Eq.~\ref{bpANN}, $\partial L/\partial O^A_k$ is the derivation of the loss function, $L$, with respect to the output of the CANN model, $O^A$, which according to Eq.~\ref{eq13}, it can be  approximated by $\partial L/\partial O^S_k$. Also, $\partial O^A_k/\partial y_d^L$ in Eq.~\ref{bpANN} is  replaced  by $\partial O^S_k/\partial y_d^L$ in Eq.~\ref{bpSNN}. Note that $O^S$ is not a function of $y^L$, hence, to approximate this derivation, with the assumption of equivalence between the firing rate of IF neurons in CSNN and activation of ReLU neurons in CANN, we simply replace the output values of $O^A$ by $O^S$ in the computations of $\partial O^A_k/\partial y_d^L$.

\begin{table*}[!htb] 
\begin{center}
\caption{Network architecture and parameter setting for Fashion-MNIST and Cifar10 datasets.}\label{tab1}
\resizebox{\textwidth}{!}{
\begin{tabular}{lllllllll}  

Dataset & Architecture  & $\theta$ & $T$& $\eta$& $\beta_1$ &$\beta_2$& $\epsilon$ &$\lambda$  \\ 
\hline
Fashion & 128C3-128C3-P2-128C3-P2-1024F-256F-10F  &  2 & 50 &$10^{-3}$ & 0.8&0.99&$10^{-7}$& $10^{-5}$\\
Cifar10& 256C3-512C3-P2-512C3-512C3-512C3-P2-256C3-1024F-512F-256F-10F  & 3 & 60 & $10^{-3} $& 0.8&0.99&$10^{-7}$& $10^{-5}$  
\end{tabular}
}
\end{center}
\end{table*}
\section{Results}
To evaluate the proposed proxy learning method, we performed experiments on two benchmark datasets of Fashion-MNIST and Cifar10. For both datasets, we used deep convolutional networks with architectural details and parameter settings provided in Table~\ref{tab1}. We use Adam optimizer with parameters $\beta_1$, $\beta_2$, and $\epsilon$ and L2-norm regularization with parameter $\lambda$. Our proxy learning  outperforms conversion and tandem learning methods by reaching 94.5\% and 92.50\% on Fashion-MNIST and Cifar10 datasets, respectively. In the following subsections, results on both datasets are provided in more details.

\subsection{Fashion-MNIST}
Fashion-MNIST~\cite{xiao2017fashion} is a fashion product image dataset with 10 classes (T-shirt, Trouser, Pullover, Dress, Coat, Sandal, Shirt, Sneaker, Bag, and Ankle boot). Images are gathered from the thumbnails of the clothing  products on an online shopping website. The Fashion-MNIST dataset contains 60,000 images of size $28\times28$ pixels as the train set. The test set contains 10,000  images ($1000$ images per class). As presented in Table~\ref{tab1}, the proposed network is comprised of three convolutional, two pooling, and three fully connected layer.

\begin{table*}[!htb]
\footnotesize
\begin{center}
\caption{Classification accuracies of different CSNNs with different learning rules on Fashion-MNIST. $T$ is the simulation time. The STDBP and STiDi-BP terms stand for spike-time-dependent error backpropagation and spike-time-displacement-based error backpropagation, respectively.}\label{tab2}
\begin{tabular}{lllcl}  

Model & Acc (\%) &  Network& $T$ & Learning \\ 
\hline
Cheng et al. (2020)~\cite{cheng2020lisnn}& 92.07& 4-layer CSNN&20& Surrogate Gradient Learning\\
Fang et al. (2021)~\cite{fang2020incorporating} & 94.36 &  6-layer CSNN &8&Surrogate Gradient Learning\\
 Yu et al. (2021)~\cite{yu2021constructing}& 92.11 &  4-layer CSNN &220&ANN-SNN Conversion\\
W. Zhang et al. (2020)~\cite{zhang2020temporal}& 92.83 & 4-layer CSNN &5&Spike Sequence Learning\\
M. Zhang et al. (2020)~\cite{zhang2020rectified} & 90.1& 5-layer CSNN&-& STDBP\\
Mirsadeghi et al. (2021)~\cite{mirsadeghi2021spike}&92.8&4-layer CSNN&100& STiDi-BP\\
\hline
This work (CANN) & 94.60 &  6-layer CANN &- & Backpropagation\\
This work (CSNN) & 84.63 &  6-layer CSNN & 50  & ANN-to-SNN Conversion\\
This work (CSNN) & 93.12 &  6-layer CSNN & 100  & ANN-to-SNN Conversion\\
This work (CSNN) & 94.50 &  6-layer CSNN & 200  & ANN-to-SNN Conversion\\
This work (CSNN) & 94.41 &  6-layer CSNN & 50  & Surrogate Gradient Learning\\
\hline
\textbf{This work (CSNN)} & \textbf{94.56} &  \textbf{6-layer CSNN} &\textbf{50}& \textbf{Proxy Learning}
\end{tabular}
\end{center}
\end{table*}

Table~\ref{tab2} provides the categorization accuracy of the proposed network with proxy learning along with the accuracy of other recent spiking neural networks on Fashion-MNIST dataset. Using a 6-layer CSNN architecture our proxy learning method could reach 94.56\% with $T=50$ accuracy and outperform other CSNNs with different learning methods. Interestingly, our network with proxy learning could surpass other networks trained with surrogate gradient learning (SGL). To do a fair comparison, we trained the same CSNN as ours using surrogate gradient learning method (with arc-tangent surrogate function~\cite{fang2020incorporating}) that reached to the best accuracy of 94.41\% with $T=50$. We also trained a CANN with the same architecture to our CSNN using backpropagation that reached to 94.60\% accuracy best (it is 0.04\% better than proxy learning). We also converted this ANN to an SNN with IF neurons (using the conversion method in~\cite{ding2021optimal}) and evaluated it on Fashion-MNIST with different simulation times. As mentioned before, conversion methods require long simulation times to reach acceptable accuracies. With  50 time steps the converted CSNN could only reach to 84.63\% accuracy and it required  more 200 time steps for 94.50\% accuracy.

\begin{figure}[t]
    \centering
    \includegraphics[scale=0.6]{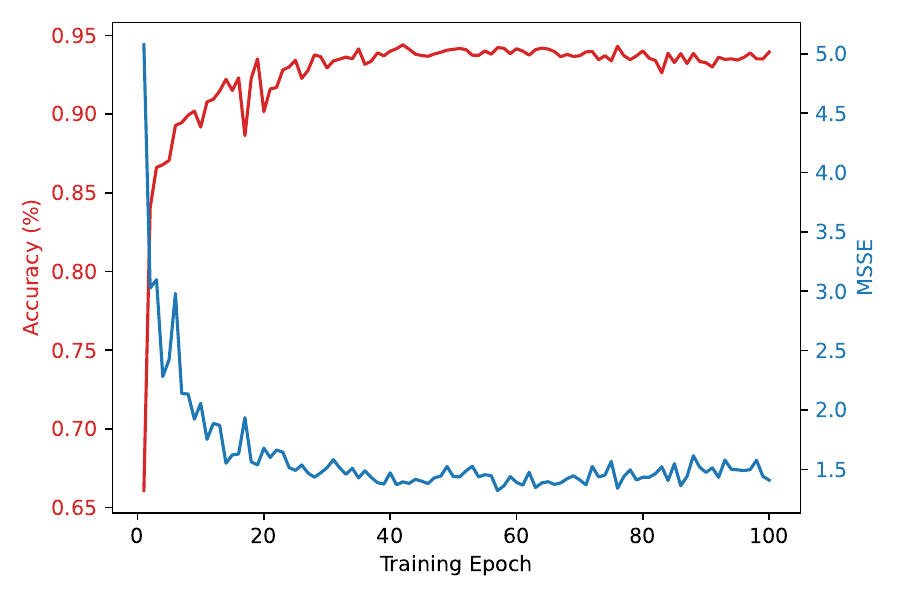}
    \caption{The categorization accuracy and  MSSE  of the proposed CSNN with proxy learning and simulation time of $T=50$ over testing set of Fashion-MNIST.}
    \label{fig:loss}
\end{figure}

The categorization accuracy and the mean sum of squared error (MSSE) of the proposed CSNN with the simulation time of $T=50$ over the test set of Fashion-MNIST dataset is provided in Fig.~\ref{fig:loss}. Only 30 epochs are enough to reach an accuracy above 94.0\% and a MSSE lower than 1.5.

\begin{figure}[t]
    \centering
    \includegraphics[scale=0.6]{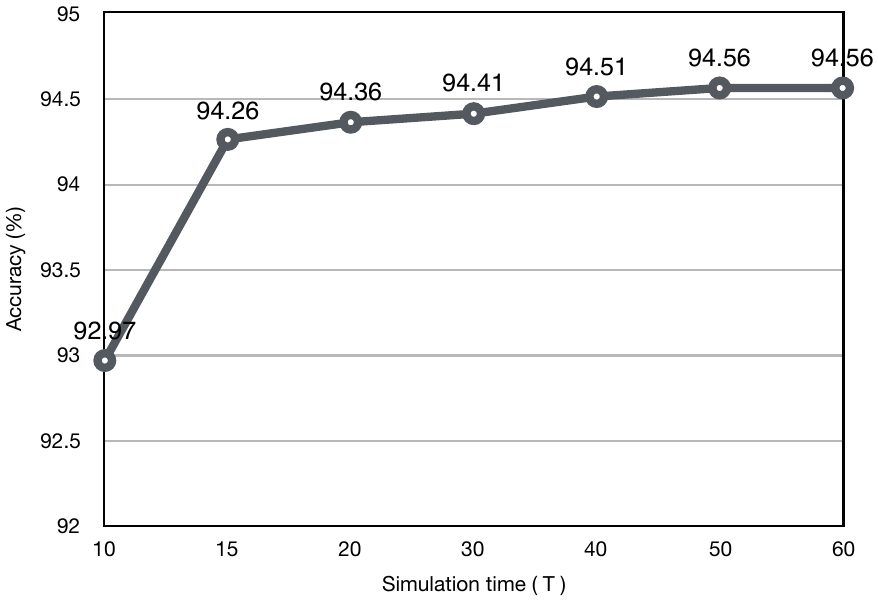}
    \caption{Classification accuracy of the proposed CSNN trained via proxy learning on Fashion-MNIST with maximum simulation time varying from $T=10$ to $T=60$. The accuracy increases by $T$ and reaches to 94.56 at $T=50$.}
    \label{fig:my_label1}
\end{figure}

As explained in section~\ref{IFrelu}, we assume that IF neurons in CSNN approximate ReLU neurons in the proxy CANN. However, it is expected that this approximation should get more accurate by increasing the maximum simulation time, $T$. To verify this expectation,  we evaluated the recognition accuracy of the CSNN model (see Fig.~\ref{fig:my_label1}) trained with proxy learning on Fashion-MNIST dataset with the simulation time varying from $T=10$ to $T=60$. As seen in Fig.~\ref{fig:my_label1}, the recognition accuracy increases as $T$ is increased. The model reached to the reasonable accuracy of 94.26\% from $T=15$ and then ascends up to 94.56\% with $T=50$. 
\begin{figure}[!htb]
    \centering
    \includegraphics[scale=0.5]{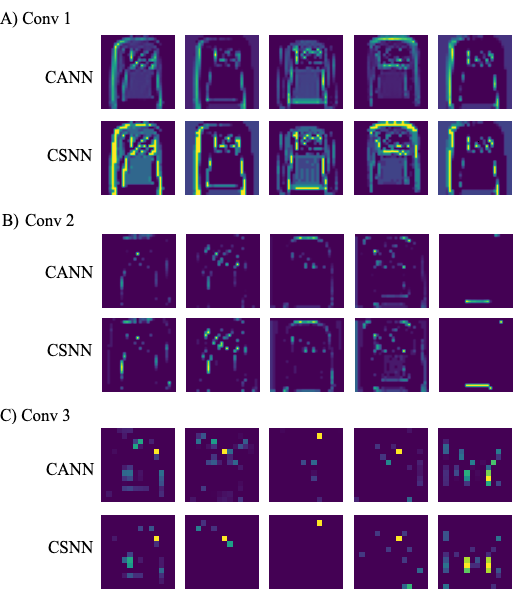}
    \caption{The output feature maps of five randomly selected filters in different convolutional layers of both CSNN and CANN networks over a randomly selected image from Fashion-MNIST. The feature maps of the CSNN network are obtained by computing the spike counts of IF neurons in each map. The firing rates of IF neurons can well approximate the activations of corresponding ReLU neurons.}
    \label{fig:my_label2}
\end{figure}

In Fig.~\ref{fig:my_label2} we plotted the output feature maps of five randomly selected convolutional filters in different layers of the CANN model and their corresponding in the CSNN model. Subfigures~\ref{fig:my_label2}A, \ref{fig:my_label2}B, and \ref{fig:my_label2}C respectively show the selected feature maps in the first, second, and third convolutional layers. In each subfigure, the top (bottom) row belongs to the feature maps of the CANN (CSNN) model. The feature maps of the CSNN model are obtained by computing the spike count (i.e., firing rate) of each IF neuron. As seen, the activation of  ReLU neurons in CANN layers is well approximated by the firing rates of corresponding IF neurons in the CSNN model.

\subsection{Cifar10}
Cifar10 is a widely-used benchmark dataset in deep learning and suitable for evaluating spiking neural network on natural image classification tasks.  Cifar10 consists color images from 10 different classes, with a standard split of 50,000 and 10,000 for training and testing, respectively. To solve Cifiar10 classification task, we developed a 10-layer CSNN trained via proxy. Architectural details of the proposed network are provided in Table~\ref{tab1}.  

The classification accuracy of the proposed network along with those of other CSNNs trained by different learning strategies including surrogate gradient learning, ANN-to-SNN conversion, and tandem learning are presented in Table~\ref{tab3}. Our proposed network could reach 93.11\% categorization accuracy on Cifar10 with $T=60$ and outperform any other CSNN trained  listed in Table~\ref{tab3}, except Fang et al. (2021)~\cite{fang2020incorporating} that use surrogate gradient in CSNNs with Leaky-IF neurons with trainable membrane time constants (i.e., each spiking neuron layer has an independent and trainable membrane time constant). Although, they reached 0.04\% better accuracy than us, implementing large CSNNs with Leaky-IF neurons having different time constants, independent of the implementation platform, is highly expensive in terms of memory and computation. However, we use simple IF neurons with no leak and no need for extra parameters, that is easy to implement and has low memory and computational costs. 
\begin{table*}[!htb] 
\footnotesize
\begin{center}
\caption{Classification accuracies of different CSNNs with different learning rules on Cifar10. $T$ is the simulation time. The STDB and SGL terms stand for spike-time-dependent backpropagation and surrogate gradient learning, respectively.}\label{tab3}
\begin{tabular}{lllcl}  

Model & Acc (\%) &  Network & $T$&Learning \\ 
\hline
Y. Wu et al. (2019)~\cite{wu2019direct}&  90.53 &8-layer CSNN &12&Surrogate Gradient Learning\\
Syed et al. (2021) ~\cite{syed2021exploring}&91.58 &VGG-13 (CSNN)&15& Surrogate Gradient Learning\\
Fang et al. (2021)~\cite{fang2020incorporating} & 93.50 &  9-layer CSNN &8&Surrogate Gradient Learning \\
Zheng et al. (2021)~\cite{zheng2020going} & 93.16 &  ResNet-19 &6&Surrogate Gradient Learning \\
Rueckauer et al. (2017)~\cite{rueckauer2017conversion} & 90.85 &VGG-16 (CSNN) &400& ANN-to-SNN conversion\\
Rathi et al. (2020)~\cite{rathi2020enabling}&92.22 & Resnet-20 (SNN)  &250& ANN-to-SNN conversion\\
Rathi et al. (2020)~\cite{rathi2020enabling}&91.13 & VGG-16 (CSNN)  &100& ANN-to-SNN conversion + STDB\\
Sengupta et al. (2019)~\cite{sengupta2019going}& 91.46 & VGG-16 (CSNN)&2500& ANN-to-SNN conversion \\
Lee et al. (2019)~\cite{lee2020enabling}& 91.59 & ResNet-11 (CSNN)&3000& ANN-to-SNN conversion\\
Liu et al. (2022)~\cite{liu2022spikeconverter} & 91.47 &  ResNet-20 &16&ANN-to-SNN conversion \\
Rathi et al. (2020)~\cite{rathi2020diet}&92.70& VGG-16 (CSNN)&5& ANN-to-SNN conversion + SGL\\
J. Wu et al. (2021)~\cite{wu2021tandem}& 90.98& CifarNet (CSNN)&8&Tandem Learning\\
J. Wu et al. (2020)~\cite{wu2020progressive}& 91.24& VGG-11 (CSNN)&16 &Progressive Tandem Learning\\
\hline
This work (CANN) & 93.20 &  10-layer CANN &-& Backpropagation\\

This work (CSNN) & 89.14&  10-layer CSNN &60 & ANN-to-SNN conversion\\
This work (CSNN) & 92.91&  10-layer CSNN & 120& ANN-to-SNN conversion\\
This work (CSNN) & 93.16&  10-layer CSNN & 240& ANN-to-SNN conversion\\

This work (CSNN) & 92.85 &  10-layer CSNN &60& Surrogate Gradient Learning\\
\hline
\textbf{This work (CSNN)} & \textbf{93.11} &  \textbf{10-layer CSNN} &\textbf{60} & \textbf{Proxy Learning}

\end{tabular}
\end{center}
\end{table*}

Interestingly, our proposed CSNN with proxy learning could significantly outperform  CSNNs with tandem learning rule~\cite{wu2021tandem,wu2020progressive}. This might be due to inconsistency between the forward and backward passes of tandem learning. In our proxy learning method, only the final output of the CANN is replaced with  that of the CSNN, and hence, the forward pass of the two networks are totally independent. However, in the forward pass of tandem learning,  CANN layers are disconnected from each other and receive the spike counts of the previous CSNN layer as their input, while in the backward pass, the CSNN error backpropagates through the CANN layers and based on their true outputs, without the intervene of CSNN layers. 

\begin{figure}[t]
    \centering
    \includegraphics[scale=0.6]{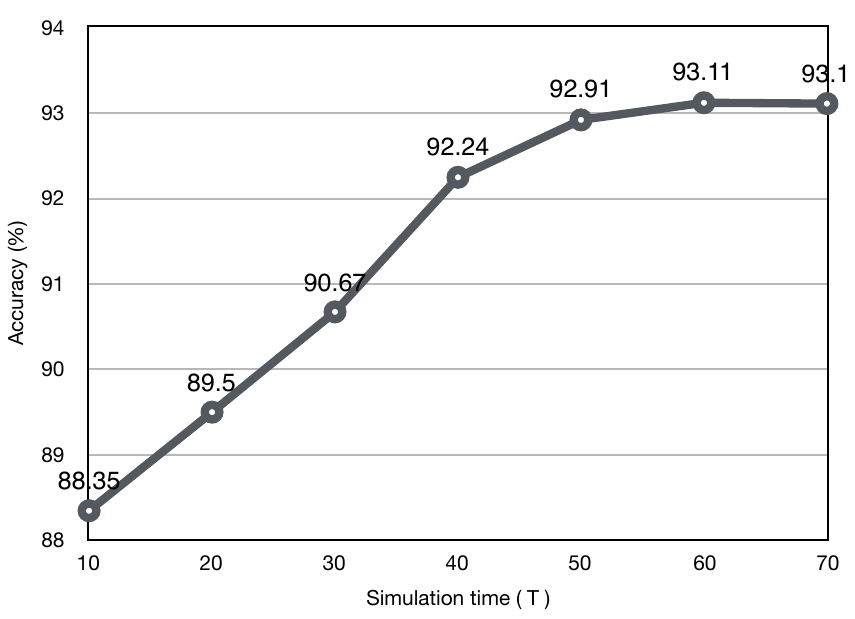}
    \caption{Classification accuracy of the proposed CSNN  trained via proxy learning on Cifar10 with maximum simulation time varying from $T=10$ to $T=70$. The accuracy increases by $T$ and reaches to 93.11\% at $T=60$.}
    \label{fig:chart2}
\end{figure}

Our proposed CSNN has also outperformed CSNNs converted from CANNs with even deeper architectures. In proxy learning, the final error is computed based on the spike counts of the output layer of the CSNN, while in conversion methods, the training phase is totally independent of the CSNN. This shows that being aware of the quantized CSNN activations (the spike counts) in our proxy learning, which are ignored by conversion methods,  can lead to CSNNs with higher classification accuracy. 

Also, we developed a CANN with ReLU neurons and same architecture to our CSNN and trained it using backpropagation with ADAM optimizer (the learning parameters and conditions were the same as the CSNN model in Table~\ref{tab1}). The CANN reached the best accuracy of 93.20\% that is only 0.09\% better than our CSNN with proxy learning. We then converted this ANN to an SNN with IF neurons using the conversion method in~\cite{ding2021optimal}. With  60 time steps, that our proposed CSNN had its best accuracy, the converted CSNN could only reach to 89.15\% accuracy and it required  240 time steps to reach 93.16\% accuracy. Again, conversion methods require long simulation times to reach reasonable accuracies. 

In the last comparison, we trained a CSNN with same architecture and learning parameters to ours using surrogate gradient learning. As presented in Table~\ref{tab3}, it reached the  classification accuracy of 92.85\% at its best. Note that contrary to other CSNNs in Table~\ref{tab3} that are trained with surrogate gradient learning, ours is consists of pure IF neurons without leakage.

In another experiment, we varied the maximum simulation time from $T=10$ to $T=70$ and evaluated the classification accuracy of the proposed CSNN. As depicted in Fig.~\ref{fig:chart2}, there is a trade-off between the simulation time and the classification accuracy. The  classification accuracy starts from $88.35\%$ with $T=10$, steadily increases with $T$, and culminates at $93.11\%$ with $T=60$.

\section{Discussion}
In recent years, deep learning in ANNs has been a revolution in the field of machine learning~\cite{lecun2015deep,mondal2021co} however researchers are still struggling to develop efficient learning algorithms for SNNs with deep architectures\cite{tavanaei2019deep}. In this paper we proposed a proxy learning method for deep convolutional spiking neural networks. The main idea is that IF neurons with rate coding can approximate the activation of ReLU neurons. To do so, first, we build a proxy CANN with ReLU neurons and the same architecture as the CSNN with IF neurons. Then, we feed the input image to both networks and the forward pass is done in both networks independently. Decision in CSNN network is made by applying a softmax on the spike counts of the output neurons. Finally, the error of the CSNN model is backpropagated in the CANN model by replacing its output with that of the CSNN.

Our proxy learning method reached 94.56\% on Fashion-MNIST and 93.11\% on Cifar10 datasets and outperformed other ANN-to-SNN conversion and tandem learning methods (see Tables~\ref{tab2} and~\ref{tab3}). The main issue with the conversion methods is neglecting the temporal nature of spiking neural networks~\cite{kheradpisheh2020temporal}. Another limitation of conversion methods is a trade-off between the inference speed and classification accuracy~\cite{wu2019direct}. To reach the optimal classification accuracy, they usually require at least several hundred of inference time steps. Tandem learning methods~\cite{wu2021tandem,wu2020progressive} could resolve these issues in conversion methods. Same as our proxy learning, they assume that IF neurons with rate-code approximate artificial neurons with ReLU activation. Hence, they connect the CANN and CSNN layers in tandem and feed the spike counts of CSNN layers (not the output of the previous CANN layer) as the input of the next CANN layer.  This breaking of the forward pass in the CANN, with approximated inputs from the CSNN layers, can attenuate the cohesion of its backward pass. This problem is solved in proxy learning by separating the forward pass of the two networks.  

Although, surrogate gradient learning is one of the best direct learning algorithms for spiking neural networks~\cite{neftci2019surrogate}, but it suffers from the challenges of backpropagation through time, especially for longer simulation times, including vanishing/exploding gradients and high computational cost and memory demand. However, in our proxy learning method we do not confront with such issues, as the backpropagation is done with the proxy CANN that is a time-free network. 

An important aspect of the proposed CSNN is the use of IF neuron model with no leak. IF neurons are pure integrators and have the simplest neuronal dynamics compared to any other spiking neuron model. For instance in leaky-IF neurons, in case of no other input spikes, at every time step, the neuron membrane potential exponentially decays with a time constant, while in pure IF neuron model, the membrane potential is updated simply by increasing or decreasing it just at the arrival of input spikes and according to the input synaptic weights. Hence, IF neurons are much simpler to be implemented on different hardware and neuromorphic platforms~\cite{oh2020hardware,liang20211}, especially, in large and deep networks.

The proposed proxy learning is based on the approximation of ReLU with rate-coded IF neurons. Rate-coding is the mostly used coding scheme in SNNs, however, other coding schemes such as temporal coding and rank-order coding are more efficient in terms of the number of spikes~\cite{mostafa2017supervised,zhang2020rectified,kheradpisheh2018stdp,mozafari2018first,mirsadeghi2021stidi,kheradpisheh2020bs4nn}. Extending the proxy learning to CSNNs with temporal coding in future studies would lead to accurate and low-cost CSNNs.

\section*{Acknowledgments}
The authors would like to thank Mr. Wei Fang who helped us in the implementation of our idea with his Spikingjelly package designed for developing deep spiking neural networks available at \url{https://github.com/fangwei123456/spikingjelly}.

\begin{thebibliography}{10}

\bibitem{thompson2020computational}
N.~C. Thompson, K.~Greenewald, K.~Lee and G.~F. Manso, The computational limits
  of deep learning, {\em arXiv preprint arXiv:2007.05558}   (2020).

\bibitem{eshraghian2021training}
J.~K. Eshraghian, M.~Ward, E.~Neftci, X.~Wang, G.~Lenz, G.~Dwivedi,
  M.~Bennamoun, D.~S. Jeong and W.~D. Lu, Training spiking neural networks
  using lessons from deep learning, {\em arXiv preprint arXiv:2109.12894}
  (2021).

\bibitem{zenke2021visualizing}
F.~Zenke, S.~M. Boht{\'e}, C.~Clopath, I.~M. Com{\c{s}}a, J.~G{\"o}ltz,
  W.~Maass, T.~Masquelier, R.~Naud, E.~O. Neftci, M.~A. Petrovici {\em et~al.},
  Visualizing a joint future of neuroscience and neuromorphic engineering, {\em
  Neuron} {\bf 109}(4)  (2021)  571--575.

\bibitem{bouvier2019spiking}
M.~Bouvier, A.~Valentian, T.~Mesquida, F.~Rummens, M.~Reyboz, E.~Vianello and
  E.~Beigne, Spiking neural networks hardware implementations and challenges: A
  survey, {\em ACM Journal on Emerging Technologies in Computing Systems
  (JETC)} {\bf 15}(2)  (2019)  1--35.

\bibitem{tavanaei2019deep}
A.~Tavanaei, M.~Ghodrati, S.~R. Kheradpisheh, T.~Masquelier and A.~Maida, Deep
  learning in spiking neural networks, {\em Neural Networks} {\bf 111}  (2019)
  47--63.

\bibitem{pfeiffer2018deep}
M.~Pfeiffer and T.~Pfeil, Deep learning with spiking neurons: opportunities and
  challenges, {\em Frontiers in neuroscience} {\bf 12}  (2018) p. 774.

\bibitem{rueckauer2018conversion}
B.~Rueckauer and S.-C. Liu, Conversion of analog to spiking neural networks
  using sparse temporal coding, {\em 2018 IEEE International Symposium on
  Circuits and Systems (ISCAS)\/},  IEEE2018, pp. 1--5.

\bibitem{rathi2020enabling}
N.~Rathi, G.~Srinivasan, P.~Panda and K.~Roy, Enabling deep spiking neural
  networks with hybrid conversion and spike timing dependent backpropagation,
  {\em arXiv preprint arXiv:2005.01807}   (2020).

\bibitem{sengupta2019going}
A.~Sengupta, Y.~Ye, R.~Wang, C.~Liu and K.~Roy, Going deeper in spiking neural
  networks: Vgg and residual architectures, {\em Frontiers in neuroscience}
  {\bf 13}  (2019) p.~95.

\bibitem{10.3389/fnins.2020.00119}
C.~Lee, S.~S. Sarwar, P.~Panda, G.~Srinivasan and K.~Roy, Enabling spike-based
  backpropagation for training deep neural network architectures, {\em
  Frontiers in Neuroscience} {\bf 14}  (2020) p. 119.

\bibitem{deng2021optimal}
S.~Deng and S.~Gu, Optimal conversion of conventional artificial neural
  networks to spiking neural networks, {\em International Conference on
  Learning Representations\/}, 2021.

\bibitem{huh2018gradient}
D.~Huh and T.~J. Sejnowski, Gradient descent for spiking neural networks, {\em
  Proceedings of the 32nd International Conference on Neural Information
  Processing Systems\/}, 2018, pp. 1440--1450.

\bibitem{neftci2019surrogate}
E.~O. Neftci, H.~Mostafa and F.~Zenke, Surrogate gradient learning in spiking
  neural networks: Bringing the power of gradient-based optimization to spiking
  neural networks, {\em IEEE Signal Processing Magazine} {\bf 36}(6)  (2019)
  51--63.

\bibitem{bohte2011error}
S.~M. Bohte, Error-backpropagation in networks of fractionally predictive
  spiking neurons, {\em International Conference on Artificial Neural
  Networks\/},  Springer2011, pp. 60--68.

\bibitem{esser2016convolutional}
S.~K. Esser, P.~A. Merolla, J.~V. Arthur, A.~S. Cassidy, R.~Appuswamy,
  A.~Andreopoulos, D.~J. Berg, J.~L. McKinstry, T.~Melano, D.~R. Barch {\em
  et~al.}, Convolutional networks for fast, energy-efficient neuromorphic
  computing, {\em Proceedings of the national academy of sciences} {\bf
  113}(41)  (2016)  11441--11446.

\bibitem{shrestha2018slayer}
S.~B. Shrestha and G.~Orchard, Slayer: spike layer error reassignment in time,
  {\em Proceedings of the 32nd International Conference on Neural Information
  Processing Systems\/}, 2018, pp. 1419--1428.

\bibitem{bellec2018long}
G.~Bellec, D.~Salaj, A.~Subramoney, R.~Legenstein and W.~Maass, Long short-term
  memory and learning-to-learn in networks of spiking neurons, {\em Advances in
  Neural Information Processing Systems\/}, 2018, pp. 787--797.

\bibitem{zimmer2019technical}
R.~Zimmer, T.~Pellegrini, S.~F. Singh and T.~Masquelier, Technical report:
  supervised training of convolutional spiking neural networks with pytorch,
  {\em arXiv preprint arXiv:1911.10124}   (2019).

\bibitem{pellegrini2021low}
T.~Pellegrini, R.~Zimmer and T.~Masquelier, Low-activity supervised
  convolutional spiking neural networks applied to speech commands recognition,
  {\em 2021 IEEE Spoken Language Technology Workshop (SLT)\/},  IEEE2021, pp.
  97--103.

\bibitem{pellegrini2021fast}
T.~Pellegrini and T.~Masquelier, Fast threshold optimization for multi-label
  audio tagging using surrogate gradient learning, {\em ICASSP 2021-2021 IEEE
  International Conference on Acoustics, Speech and Signal Processing
  (ICASSP)\/},  IEEE2021, pp. 651--655.

\bibitem{kheradpisheh2020temporal}
S.~R. Kheradpisheh and T.~Masquelier, Temporal backpropagation for spiking
  neural networks with one spike per neuron, {\em International Journal of
  Neural Systems} {\bf 30}(06)  (2020) p. 2050027.

\bibitem{zhang2020spike}
M.~Zhang, J.~Wang, Z.~Zhang, A.~Belatreche, J.~Wu, Y.~Chua, H.~Qu and H.~Li,
  Spike-timing-dependent back propagation in deep spiking neural networks, {\em
  arXiv preprint arXiv:2003.11837}   (2020).

\bibitem{sakemi2021supervised}
Y.~Sakemi, K.~Morino, T.~Morie and K.~Aihara, A supervised learning algorithm
  for multilayer spiking neural networks based on temporal coding toward
  energy-efficient vlsi processor design, {\em IEEE Transactions on Neural
  Networks and Learning Systems}   (2021).

\bibitem{zhang2020temporal}
W.~Zhang and P.~Li, Temporal spike sequence learning via backpropagation for
  deep spiking neural networks, {\em Advances in Neural Information Processing
  Systems} {\bf 33}  (2020).

\bibitem{bohte2002error}
S.~M. Bohte, J.~N. Kok and H.~La~Poutre, Error-backpropagation in temporally
  encoded networks of spiking neurons, {\em Neurocomputing} {\bf 48}(1-4)
  (2002)  17--37.

\bibitem{zhou2019direct}
S.~Zhou, Y.~Chen, Q.~Ye and J.~Li, Direct training based spiking convolutional
  neural networks for object recognition, {\em arXiv preprint arXiv:1909.10837}
    (2019).

\bibitem{wunderlich2021event}
T.~C. Wunderlich and C.~Pehle, Event-based backpropagation can compute exact
  gradients for spiking neural networks, {\em Scientific Reports} {\bf 11}(1)
  (2021)  1--17.

\bibitem{mostafa2017supervised}
H.~Mostafa, Supervised learning based on temporal coding in spiking neural
  networks, {\em IEEE transactions on neural networks and learning systems}
  {\bf 29}(7)  (2017)  3227--3235.

\bibitem{wu2021tandem}
J.~Wu, Y.~Chua, M.~Zhang, G.~Li, H.~Li and K.~C. Tan, A tandem learning rule
  for effective training and rapid inference of deep spiking neural networks,
  {\em IEEE Transactions on Neural Networks and Learning Systems}   (2021).

\bibitem{wu2020progressive}
J.~Wu, C.~Xu, D.~Zhou, H.~Li and K.~C. Tan, Progressive tandem learning for
  pattern recognition with deep spiking neural networks, {\em arXiv preprint
  arXiv:2007.01204}   (2020).

\bibitem{fang2020incorporating}
W.~Fang, Z.~Yu, Y.~Chen, T.~Masquelier, T.~Huang and Y.~Tian, Incorporating
  learnable membrane time constant to enhance learning of spiking neural
  networks, {\em International Conference on Computer Vision (ICCV)\/}, 2021.

\bibitem{rueckauer2017conversion}
B.~Rueckauer, I.-A. Lungu, Y.~Hu, M.~Pfeiffer and S.-C. Liu, Conversion of
  continuous-valued deep networks to efficient event-driven networks for image
  classification, {\em Frontiers in neuroscience} {\bf 11}  (2017) p. 682.

\bibitem{tavanaei2019bp}
A.~Tavanaei and A.~Maida, Bp-stdp: Approximating backpropagation using spike
  timing dependent plasticity, {\em Neurocomputing} {\bf 330}  (2019)  39--47.

\bibitem{xiao2017fashion}
H.~Xiao, K.~Rasul and R.~Vollgraf, Fashion-mnist: a novel image dataset for
  benchmarking machine learning algorithms, {\em arXiv preprint
  arXiv:1708.07747}   (2017).

\bibitem{cheng2020lisnn}
X.~Cheng, Y.~Hao, J.~Xu and B.~Xu, Lisnn: Improving spiking neural networks
  with lateral interactions for robust object recognition., {\em IJCAI\/},
  2020, pp. 1519--1525.

\bibitem{yu2021constructing}
Q.~Yu, C.~Ma, S.~Song, G.~Zhang, J.~Dang and K.~C. Tan, Constructing accurate
  and efficient deep spiking neural networks with double-threshold and
  augmented schemes, {\em IEEE Transactions on Neural Networks and Learning
  Systems}   (2021).

\bibitem{zhang2020rectified}
M.~Zhang, J.~Wang, B.~Amornpaisannon, Z.~Zhang, V.~Miriyala, A.~Belatreche,
  H.~Qu, J.~Wu, Y.~Chua, T.~E. Carlson {\em et~al.}, Rectified linear
  postsynaptic potential function for backpropagation in deep spiking neural
  networks, {\em arXiv preprint arXiv:2003.11837}   (2020).

\bibitem{mirsadeghi2021spike}
M.~Mirsadeghi, M.~Shalchian, S.~R. Kheradpisheh and T.~Masquelier, Spike time
  displacement based error backpropagation in convolutional spiking neural
  networks, {\em arXiv preprint arXiv:2108.13621}   (2021).

\bibitem{ding2021optimal}
J.~Ding, Z.~Yu, Y.~Tian and T.~Huang, Optimal ann-snn conversion for fast and
  accurate inference in deep spiking neural networks, {\em arXiv preprint
  arXiv:2105.11654}   (2021).

\bibitem{wu2019direct}
Y.~Wu, L.~Deng, G.~Li, J.~Zhu, Y.~Xie and L.~Shi, Direct training for spiking
  neural networks: Faster, larger, better, {\em Proceedings of the AAAI
  Conference on Artificial Intelligence\/},   {\bf 33}(01)2019, pp. 1311--1318.

\bibitem{syed2021exploring}
T.~Syed, V.~Kakani, X.~Cui and H.~Kim, Exploring optimized spiking neural
  network architectures for classification tasks on embedded platforms, {\em
  Sensors} {\bf 21}(9)  (2021) p. 3240.

\bibitem{zheng2020going}
H.~Zheng, Y.~Wu, L.~Deng, Y.~Hu and G.~Li, Going deeper with directly-trained
  larger spiking neural networks, {\em Proceedings of the AAAI Conference on
  Artificial Intelligence\/}, 2021.

\bibitem{lee2020enabling}
C.~Lee, S.~S. Sarwar, P.~Panda, G.~Srinivasan and K.~Roy, Enabling spike-based
  backpropagation for training deep neural network architectures, {\em
  Frontiers in neuroscience} {\bf 14}  (2020) p. 119.

\bibitem{liu2022spikeconverter}
F.~Liu, W.~Zhao, Y.~Chen, Z.~Wang and L.~Jiang, Spikeconverter: An efficient
  conversion framework zipping the gap between artificial neural networks and
  spiking neural networks, {\em Proceedings of the AAAI Conference on
  Artificial Intelligence\/}, 2022.

\bibitem{rathi2020diet}
N.~Rathi and K.~Roy, Diet-snn: Direct input encoding with leakage and threshold
  optimization in deep spiking neural networks, {\em arXiv preprint
  arXiv:2008.03658}   (2020).

\bibitem{lecun2015deep}
Y.~LeCun, Y.~Bengio and G.~Hinton, Deep learning, {\em nature} {\bf 521}(7553)
  (2015)  436--444.

\bibitem{mondal2021co}
M.~R.~H. Mondal, S.~Bharati and P.~Podder, Co-irv2: Optimized inceptionresnetv2
  for covid-19 detection from chest ct images, {\em PLoS One} {\bf 16}(10)
  (2021) p. e0259179.

\bibitem{oh2020hardware}
S.~Oh, D.~Kwon, G.~Yeom, W.-M. Kang, S.~Lee, S.~Y. Woo, J.~S. Kim, M.~K. Park
  and J.-H. Lee, Hardware implementation of spiking neural networks using
  time-to-first-spike encoding, {\em arXiv preprint arXiv:2006.05033}   (2020).

\bibitem{liang20211}
M.~Liang, J.~Zhang and H.~Chen, A 1.13 $\mu$j/classification spiking neural
  network accelerator with a single-spike neuron model and sparse weights, {\em
  2021 IEEE International Symposium on Circuits and Systems (ISCAS)\/},
  IEEE2021, pp. 1--5.

\bibitem{kheradpisheh2018stdp}
S.~R. Kheradpisheh, M.~Ganjtabesh, S.~J. Thorpe and T.~Masquelier, Stdp-based
  spiking deep convolutional neural networks for object recognition, {\em
  Neural Networks} {\bf 99}  (2018)  56--67.

\bibitem{mozafari2018first}
M.~Mozafari, S.~R. Kheradpisheh, T.~Masquelier, A.~Nowzari-Dalini and
  M.~Ganjtabesh, First-spike-based visual categorization using reward-modulated
  stdp, {\em IEEE transactions on neural networks and learning systems} {\bf
  29}(12)  (2018)  6178--6190.

\bibitem{mirsadeghi2021stidi}
M.~Mirsadeghi, M.~Shalchian, S.~R. Kheradpisheh and T.~Masquelier, Stidi-bp:
  Spike time displacement based error backpropagation in multilayer spiking
  neural networks, {\em Neurocomputing} {\bf 427}  (2021)  131--140.

\bibitem{kheradpisheh2020bs4nn}
S.~R. Kheradpisheh, M.~Mirsadeghi and T.~Masquelier, Bs4nn: Binarized spiking
  neural networks with temporal coding and learning, {\em arXiv preprint
  arXiv:2007.04039}   (2020).

\end{thebibliography}

\end{document}